\documentclass{article}

\usepackage{arxiv}

\usepackage[utf8]{inputenc} 
\usepackage[T1]{fontenc}    
\usepackage{url}            
\usepackage{booktabs}       
\usepackage{amsmath}
\usepackage{amsfonts}
\usepackage{pifont}
\usepackage{bm}
\usepackage{bbold}
\usepackage{amssymb}
\usepackage[group-separator={,}]{siunitx}
\usepackage{wrapfig,booktabs}
\usepackage{nicefrac}       
\usepackage{microtype}      
\usepackage{hyperref}       
\usepackage[nameinlink]{cleveref}
\usepackage{graphicx}
\usepackage{caption}
\usepackage{subcaption}
\usepackage{mwe}
\usepackage{apacite}
\let\cite\shortcite
\def\eqref#1{equation~\ref{#1}}

\def\1{\bm{1}}

\def\ve{{\bm{e}}}

\def\vh{{\bm{h}}}

\def\vr{{\bm{r}}}

\def\vt{{\bm{t}}}

\def\vw{{\bm{w}}}

\DeclareMathAlphabet{\mathsfit}{\encodingdefault}{\sfdefault}{m}{sl}
\SetMathAlphabet{\mathsfit}{bold}{\encodingdefault}{\sfdefault}{bx}{n}

\newcommand{\R}{\mathbb{R}}
\newcommand{\C}{\mathbb{C}}

\newcommand{\normlone}{L^1}
\newcommand{\normltwo}{L^2}
\newcommand{\normlthree}{L^3}

\newcommand{\norm}[1]{\left\lVert#1\right\rVert}

\newcommand{\cmark}{\ding{51}}
\newcommand{\xmark}{\ding{55}}

\title{BESS: Balanced Entity Sampling and Sharing for Large-Scale Knowledge Graph Completion}

\renewcommand{\undertitle}{1st Place OGB-LSC 2022 (WikiKG90Mv2 Track)}
\renewcommand{\shorttitle}{BESS: Balanced Entity Sampling and Sharing for Large-Scale Knowledge Graph Completion}

\author{
    Alberto Cattaneo\thanks{Equal contribution}\\
    Graphcore\\
    \texttt{albertoc@graphcore.ai} 
    \And 
    Daniel Justus\footnotemark[1]\\
    Graphcore\\
    \texttt{danielj@graphcore.ai}
    \And
    Harry Mellor\footnotemark[1]\\
    Graphcore\\
    \texttt{harrym@graphcore.ai}
    \And 
    Douglas Orr\footnotemark[1]\\
    Graphcore\\
    \texttt{douglaso@graphcore.ai}
    \And 
    Jerome Maloberti\\
    Graphcore\\
    \And 
    Zhenying Liu\\
    Graphcore\\
    \And 
    Thorin Farnsworth\\
    Graphcore\\
    \And 
    Andrew Fitzgibbon\\
    Graphcore\\
    \texttt{awf@graphcore.ai}
    \And 
    Blazej Banaszewski\\
    Graphcore\\
    \texttt{blazejb@graphcore.ai}
    \And 
    Carlo Luschi\\
    Graphcore\\
    \texttt{carlo@graphcore.ai}
}
\date{November 2022}

\begin{document}

\maketitle
\vspace{5mm}

\begin{abstract}
We present the award-winning submission to the WikiKG90Mv2 track of OGB-LSC@NeurIPS 2022. The task is link-prediction on the large-scale knowledge graph WikiKG90Mv2, consisting of 90M+ nodes and 600M+ edges. Our solution uses a diverse ensemble of $85$ Knowledge Graph Embedding models combining five different scoring functions (TransE, TransH, RotatE, DistMult, ComplEx) and two different loss functions (log-sigmoid, sampled softmax cross-entropy). Each individual model is trained in parallel on a Graphcore Bow~Pod$_{16}$ using BESS (Balanced Entity Sampling and Sharing), a new distribution framework for KGE training and inference based on balanced collective communications between workers. Our final model achieves a validation MRR of 0.2922 and a test-challenge MRR of 0.2562, winning the first place in the competition. The code is publicly available at: \url{https://github.com/graphcore/distributed-kge-poplar/tree/2022-ogb-submission}.
\end{abstract}

\section{Introduction}

Knowledge Graphs encode a knowledge base in the form of a heterogeneous directed graph, where facts are subject-predicate-object triples which are represented as labelled edges (relations) connecting pairs of nodes (entities). Over the past decades they have attracted growing interest, finding a wide variety of commercial applications ranging from drug discovery \cite{KG_drug_discovery} to question-answering \cite{end_to_end_question_answering} and recommender systems \cite{recommender_systems}. Knowledge Graph Embedding (KGE) models perform reasoning on knowledge graphs by learning a semantic-aware mapping of entities and relations to low-dimensional vector spaces $V_e, V_r$ respectively, such that the plausibility of triples is measured by a scoring function of the head, relation and tail embeddings $f: V_e \times V_r \times V_e \rightarrow \R$. The learned embeddings can then be used to infer missing links in the graph (Knowledge Graph Completion) and for downstream tasks. 

While the majority of the literature on KGE models focuses on relatively small graphs, real-world applications of commercial value increasingly require reasoning on graphs with hundreds of millions, or even billions, of entities and edges \cite{Wikidata,freebase}. It has therefore become paramount to investigate models with good scaling capabilities and develop effective distributed training frameworks running on multiple devices \cite{pytorch_biggraph,dglke}. KGE models are characterised by large memory requirements for storing parameters (almost entirely concentrated in the embedding tables) with sparse memory access patterns, since at each training step only the embeddings of entities and relations in the mini-batch need to be accessed and updated. This makes parallelisation of KGE models potentially challenging, as communications between workers need to be carefully managed in order to keep embeddings synchronised without incurring in excessive overheads.

The \emph{Open Graph Benchmark Large-Scale Challenge} (OGB-LSC) \cite{OGBLSC} aims to encourage the graph ML research community to work with realistically sized datasets and develop solutions able to meet real-world needs, by providing learning tasks with immediate applications on graphs at an unprecedented scale. The WikiKG90Mv2 track of the competition requires performing Knowledge Graph Completion on a graph with more than 90M entities.  Our winning solution consists of an ensemble of $85$ KGE models combining a variety of well-established scoring functions \cite{TransE,TransH,RotatE,DistMult,ComplEx}, implemented on the distributed processing framework BESS powered by Graphcore's Poplar SDK \cite{poplar} which allows for fast, communication-efficient training and inference (see \Cref{sec:acceleration}). 

\section{Task and Dataset Description} \label{sec: dataset}

\setlength\intextsep{3pt}
\begin{wraptable}{r}{6cm}
\caption{WikiKG90Mv2 dataset}\label{table: wikikg_stats}
\begin{tabular}{lr} \toprule
$\left| \mathcal{E} \right|$ & $\num{91230610}$\\
$\left| \mathcal{R} \right|$ & $\num{1387}$\\
$\left| \mathcal{T} \right|$ & $\num{601062811}$\\
\# validation queries & $\num{15000}$\\
\# test-dev queries & $\num{10000}$\\
\# test-challenge queries & $\num{10000}$\\
\bottomrule
\end{tabular}
\end{wraptable}

The WikiKG90Mv2 dataset \cite{OGBLSC} is a knowledge graph constructed from the Wikidata open knowledge base \cite{Wikidata}. We denote by $\mathcal{E}$ the set of entities (Wikidata items) in the knowledge graph and by $\mathcal{R}$ the set of relations (Wikidata linking properties). A subject-predicate-object claim is then abstracted as a triple $(h, r, t)$ with $h,t \in \mathcal{E}$ and $r \in \mathcal{R}$. The training set $\mathcal{T}$ consists of positive triples representing true facts in the knowledge base. A $768$-dimensional feature embedding vector is also provided for each entity and relation, obtained by encoding the title and description of the corresponding Wikidata entry with MPNet \cite{MPNet}.

The task is to impute missing links in the knowledge graph, by predicting the top-$10$ tail entities $t$ which are most likely to complete a query $(h, r, ?)$. The metric used is the Mean Reciprocal Rank (MRR) of the ground-truth tail among the top-$10$ candidates (with a reciprocal rank of $0$ if the ground-truth is not present in the set of predictions). The validation and test sets are extracted from snapshots of the Wikidata knowledge base at later time-stamps.

\subsection{Dataset Exploration} \label{sec:dataset}

\begin{figure}[htb]
    \centering
    \begin{subfigure}[t]{0.49\textwidth}
        \centering
        \includegraphics[height=4.3cm]{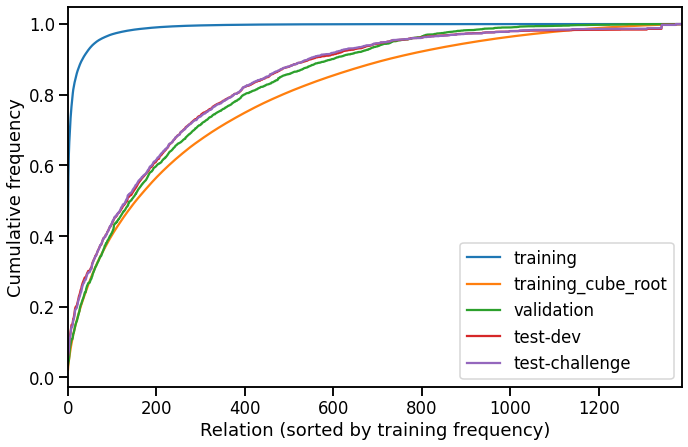}
        \subcaption{}
        \label{subfig:relation-distr}
    \end{subfigure}
    \begin{subfigure}[t]{0.49\textwidth}
        \centering
        \includegraphics[height=4.3cm]{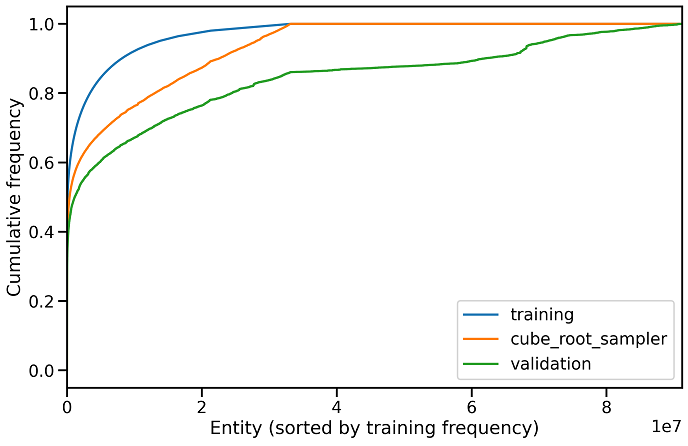}
        \subcaption{}
        \label{subfig:tail-distr}
    \end{subfigure}
    \caption{(a) Cumulative frequencies of relations in the four dataset splits. The distribution of the cube root of relation counts in the training set is also displayed. (b) Cumulative frequencies of tail entities in the training and validation sets.}
\end{figure}

Given the large number of nodes and edges in the knowledge graph, it is useful to compare statistics for the four dataset splits (training, validation, test-dev and test-challenge sets). \Cref{subfig:relation-distr} highlights a striking discrepancy in the the distribution of relations between training and validation/test sets. While the same relation (ID 481) is the one appearing most frequently in all sets, it spans more than $40\%$ of the training triples but only $3.3\%$, $5.8\%$ and $5.9\%$ of the triples in the validation, test-dev and test-challenge sets respectively. As detailed in the dataset documentation \cite{OGBLSC_WikiKG_site}, the validation and tests sets have been sampled so that the final relation counts are proportional to the cube root of the counts in the respective Wikidata dumps. Sampling from the training set with a similar strategy produces a better distribution alignment (\Cref{subfig:relation-distr}). When looking at the distribution of entities, we notice that only $\num{33179325}$ of them (roughly one third of $\left| \mathcal{E} \right|$) appear as tails in the training set. As shown in \Cref{subfig:tail-distr}, the cube root sampling strategy helps to mitigate the difference between the training and validation distributions of tails, however almost 20\% of tail entities in the validation set are never used as tails in the training set.

\section{Methodology}

\subsection{Model Architecture}

\paragraph{Encoder}
All KGE models in the final ensemble share the same shallow encoding strategy, which we describe in this paragraph. For an entity $e \in \mathcal{E}$, we denote by $\ve_F \in \R^{768}$ its MPNet text features provided in the dataset and define a trainable entity embedding $\ve_S \in \R^d$. We use linear layers $\mathbf{M}_H, \mathbf{M}_T \in \R^{d \times 768}$ to project $\ve_F$ to $\R^d$ for head and tail entities respectively, optionally with $\mathbf{M}_H = \mathbf{M}_T$. The final entity embedding is given by:
\begin{equation}
\label{eqn: embedding} 
\begin{split} 
\ve = \ve_S + \mathbf{M}_H\ve_F \qquad &\text{for head entities;}\\
\ve = \ve_S + \mathbf{M}_T\ve_F \qquad &\text{for tail entities.}
\end{split}
\end{equation}

Since the number of relations is small we do not make use of their text features, but only train a shallow embedding $\vr \in \R^k$ for each $r \in \mathcal{R}$, where $k=d/2$ for RotatE and $k=d$ otherwise.

\paragraph{Scoring Functions}
The model's decoder assigns to each triple $(h,r,t)$ a score $f(\vh, \vr, \vt) \in \R$, where $\vh, \vr, \vt$ are the embedding vectors for the head entity $h$, relation $r$ and tail entity $t$ respectively, obtained through the encoder as in \cref{eqn: embedding}. We consider five different scoring functions: TransE \cite{TransE}, TransH \cite{TransH}, RotatE \cite{RotatE}, DistMult \cite{DistMult} and ComplEx \cite{ComplEx} (see \Cref{table: decoders}). For the three distance-based scoring functions, namely TransE, TransH and RotatE, we test both $\normlone$ and $\normltwo$ distances. In the case of TransH, for each relation $r \in \mathcal{R}$ we have the additional trainable parameter given by $\vw_{r} \in \R^d$, which represents the unit normal vector to the relation-specific hyperplane onto which the entity embeddings are projected.

\begin{table}[b]
\centering
\caption{Scoring functions and their ability to model four fundamental relation properties: S = Symmetry; AS = Antisymmetry; I = Inversion; C = Composition. For RotatE and ComplEx we assume $d$ even and denote by $\C^{\frac{d}{2}}$ the vector space $\mathbb{R}^d = (\mathbb{R} \oplus i\mathbb{R})^{\frac{d}{2}}$ with the structure of $\R$-algebra induced by the product of complex numbers. $\circ$ denotes the Hadamard product; $p \in \left\{ 1,2 \right\}$.}\label{table: decoders}
\begin{tabular}{lllcccc}
\toprule
\textbf{Model} & \multicolumn{2}{c}{\textbf{Scoring function}} & \textbf{S} & \textbf{AS} & \textbf{I} & \textbf{C} \\
\midrule
TransE & $-\norm{\vh + \vr - \vt}_p$ & $\vh, \vr, \vt \in \R^d$ & \xmark & \cmark & \cmark & \cmark \\
TransH & $-\norm{\left( \vh - \vw_{r}^\intercal \vh \vw_{r} \right) + \vr - \left(\vt - \vw_{r}^\intercal \vt \vw_{r} \right)}_p$ & $\vh, \vr, \vw_{r}, \vt \in \R^d$ & \cmark & \cmark & \xmark & \xmark \\
RotatE & $-\norm{\vh \circ e^{i \vr} - \vt}_p$ & $\vh, \vt \in \C^{\frac{d}{2}}, \vr \in \R^{\frac{d}{2}}$ & \cmark & \cmark & \cmark & \cmark \\
DistMult & $\langle \vr, \vh, \vt \rangle$ & $\vh, \vr, \vt \in \R^d$ & \cmark & \xmark & \xmark & \xmark \\
ComplEx & $\mathrm{Re}\langle \vr, \vh, \overline{\vt} \rangle$ & $\vh, \vr, \vt \in \C^{\frac{d}{2}}$ & \cmark & \cmark & \cmark & \xmark \\
\bottomrule
\end{tabular}
\end{table}

\paragraph{Loss Functions}
Following standard convention, we optimise KGE models by imposing that the score $f(\vh, \vr, \vt)$ of a positive triple $(h,r,t) \in \mathcal{T}$ is larger than the score of (pseudo)negative samples $(h,r,t'_i), i=1,\dots,N$, obtained by randomly replacing the tail entity $t$. Two different loss functions $\mathcal{L}$ are considered.
\begin{itemize}
    \item \textbf{Log-sigmoid loss} \cite{RotatE}. 
    $$ \mathcal{L}(h,r,t) = -\log \sigma(\gamma + f(\vh, \vr, \vt)) - \sum_{i=1}^N w_i \log \sigma(-\gamma - f(\vh, \vr, \vt'_i))$$
    \noindent where $\gamma > 0$ is a fixed margin for distance-based scoring functions ($\gamma = 0$ for DistMult and ComplEx), $\sigma$ is the sigmoid function and we use self-adversarial negative sample weighting
    $$ w_i = \texttt{StopGrad}\left( \frac{e^{a \cdot f(\vh, \vr, \vt'_i)}}{\sum_{j=1}^N e^{a \cdot f(\vh, \vr, \vt'_j)}} \right)$$
    to upweight negative samples with higher scores (i.e.\ those which are more difficult for the current model to tell apart). Here $a \ge 0$ is a hyperparameter tuning the temperature of self-adversarial negative sampling.
    \item \textbf{Sampled softmax cross entropy loss} \cite{large_vocabulary}. A variant of plain softmax cross entropy loss which uses the target class and a set of $N$ negative samples to estimate the log-sum-exp of logits over all possible classes (in our case, the 90M+ entities in the knowledge graph). We can lower the variance of such estimator by separating the contribution to the log-sum-exp of the target class and introducing a correction $c = \log\frac{|\mathcal{E}| -1}{N}$ for the other terms as follows:
    $$ \mathcal{L}(h,r,t) = -f(\vh, \vr, \vt)  +  \log \left( e^{f(\vh, \vr, \vt)} + \sum_{i=1}^N e^{f(\vh, \vr, \vt'_i) + c}\right).$$
\end{itemize}

\paragraph{Regularisation}
We regularise both losses with the $\normlthree$ norm of embedding vectors. This was motivated in \shortcite{lp_reg} for tensor-decomposition scoring functions such as DistMult and ComplEx, however we find beneficial effects also with distance-based scores. We compute the $\normlthree$ norm on the final entity embedding in \cref{eqn: embedding} and its separate components, namely the trainable shallow embedding and the text feature projection. For a micro-batch $\mathcal{B}$ with (shared) negative tails $t'_i, i=1,\dots,N$, the regularisation term added to the micro-batch loss is $\lambda_T\Omega^\mathcal{B}_T +  \lambda_S\Omega^\mathcal{B}_S + \lambda_F\Omega^\mathcal{B}_F$, where $\lambda_T, \lambda_S, \lambda_F$ are distinct regularisation parameters and
\begin{align*}
\Omega^\mathcal{B}_T &=  \sum_{(h,r,t) \in \mathcal{B}} \left( \norm{\vh}_3 + \norm{\vt}_3 \right) + \sum_{i=1}^N \norm{\vt'_i}_3,\\
\Omega^\mathcal{B}_S &=  \sum_{(h,r,t) \in \mathcal{B}} \left( \norm{\vh_{S}}_3 + \norm{\vt_{S}}_3 \right) + \sum_{i=1}^N \norm{\vt'_{S,i}}_3,\\
\Omega^\mathcal{B}_F &=   \sum_{(h,r,t) \in \mathcal{B}} \left( \norm{\mathbf{M}_H \vh_{F}}_3 + \norm{\mathbf{M}_T \vt_{F}}_3 \right) + \sum_{i=1}^N \norm{\mathbf{M}_T \vt'_{F,i}}_3.
\end{align*}

As an additional form of regularisation, we also experiment with applying dropout \cite{dropout} to the linear projections $\mathbf{M}_H\ve_F$,  $\mathbf{M}_T\ve_F$ in \cref{eqn: embedding}, before summing the output with $\ve_S$.

\subsection{Inference}
Given a test query $(h, r, ?)$, inference is performed by traversing all entities in the knowledge graph and selecting the tails $t$ that realise the top-$K$ scores in $\left\{ f(\vh, \vr, \vt), t \in \mathcal{E} \right\}$. Despite this approach having time complexity $O(|\mathcal{E}| + |\mathcal{E}|\log K)$, our BESS distributed setup allows us to perform validation and testing fast enough to avoid any form of bias introduced by candidate selection methods \cite{PIE}.

\paragraph{Ensemble}
We use a power-rank ensembling strategy generalising \shortciteA{reciprocal_rank_fusion} to combine the predictions of $M$ trained individual models for a query $(h,r,?)$. Let $\left\{ t^m_1, \dots, t^m_K \right\}$ be the top-$K$ ranked set of tails predicted by the $m$-th model, for $K \geq 10$. For a power hyperparameter $p \neq 0$, we assign the following rank-based score to each entity $t \in \mathcal{E}$:
\begin{equation}
    \label{eqn: ensemble}
    s(t) = \sum_{m=1}^M s_m(t), \qquad s_m(t) = 
    \begin{cases} 
    - \mathop{\mathrm{sgn}}(p) k^p \quad &\text{if } t = t^m_k \\ 
    - \mathbb{1}_{p>0} \cdot (K + 1)^p \quad &\text{if } t \notin \left\{ t^m_1, \dots, t^m_K \right\} 
    \end{cases}
\end{equation}
\noindent and select the entities with top-10 $s(t)$ values as final ranked predictions.

\section{Acceleration and Distribution Strategy} \label{sec:acceleration}

The distribution scheme BESS (Balanced Entity Sampling and Sharing) involves a master process coordinating $D$ workers (in our case, single Graphcore IPUs), with the key feature that workers can exchange data directly between them via collective communications, so that no additional parameter server is needed. 

\paragraph{Partitioning} 
Each embedding table is randomly partitioned row-wise across the $D$ workers, in shards of equal sizes stored in the workers' memory. We denote by $\mathcal{E}_1, \dots, \mathcal{E}_D$ the partitions of the set of entities $\mathcal{E}$, with $|\mathcal{E}_i| = \left\lceil \frac{|\mathcal{E}|}{D} \right\rceil$. This splitting induces a partitioning of the triples $(h,r,t)$ in the knowledge graph based on the location of the head and tail entities: $\mathcal{T}_{i,j} = \left\{ (h,r,t) \in \mathcal{T}: h \in \mathcal{E}_i, t \in \mathcal{E}_j \right\}$, $i,j = 1, \dots, D$. Even with random partitioning of entities, the size variance of the $D^2$ partitions $\mathcal{T}_{i,j}$ will depend on the connectivity patterns of the specific knowledge graph; for WikiKG90Mv2 we find them to be always sufficiently well-balanced.

Since the number of relations in knowledge graphs is typically small (compared to the number of entities), we can afford to use an AllGather collective to reconstruct the whole relation embedding table on each worker before extracting the relevant embeddings needed to compute the micro-batch loss or predictions. The same strategy is used to split and retrieve  the head and tail feature projection matrices $\mathbf{M}_H, \mathbf{M}_T$ and all weights other than entity embeddings, whose sharing requires an ad hoc strategy that is different for training and inference.

\paragraph{Training}
At each training step, the master process samples a micro-batch $\mathcal{B}_i$ for each of the workers, with $|\mathcal{B}_i| = B \equiv 0 \pmod{D}$ for $i=1,\dots,D$. All triples $(h,r,t) \in \mathcal{B}_i$ have $h \in \mathcal{E}_i$, while tail entities are equally distributed across partitions $\mathcal{E}_j$, i.e.\
\begin{equation} \label{eqn: microbatch}
\mathcal{B}_i = \bigcup_{j=1}^D \mathcal{B}_{i,j}, \qquad \mathcal{B}_{i,j} \subset \mathcal{T}_{i,j}, \; |\mathcal{B}_{i,j}| = \frac{B}{D}.
\end{equation}

Triples $(h,r,t) \in \mathcal{B}_{i,j}$ are sampled (with replacement) from $\mathcal{T}_{i,j}$ according to the following probability distribution:
\begin{equation*}
p((h,r,t)) = p((h,r,t) | r) p(r) = \frac{1}{n_r} \frac{\sqrt[3]{n_r}}{\sum_{r' \in \mathcal{R}} \sqrt[3]{n_{r'}} }
\end{equation*}
\noindent where, for a relation $r' \in \mathcal{R}$, we denote $n_{r'} := \left| \left\{(h,r,t) \in \mathcal{T}_{i,j} : r = r' \right\} \right|$. As motivated in \Cref{sec:dataset}, we force the distribution of relations produced by the sampler to be proportional to the cube root of relation frequencies in the training set, in order to better align it with the validation and test sets, thus reducing distribution shift.

Together with positive triples, the master process also samples sets of entities to construct negative samples. We adopt \textit{negative sample sharing}, i.e.\ use the same set of corrupted tails for all triples in a micro-batch. This allows us to increase the effective negative sample size without increasing communication costs, while also reducing the computational cost of scoring negative triples (as negative tail embeddings can be broadcasted across the micro-batch). The set of negative tails used for the micro-batch $\mathcal{B}_{i}$ is given by
\begin{equation} \label{eqn: negsamples} 
\mathcal{N}_i =  \bigcup_{j=1}^D \mathcal{N}_{i,j}, \qquad \mathcal{N}_{i,j} \subset \mathcal{E}_j, \; |\mathcal{N}_{i,j}| = \frac{N}{D}
\end{equation}
\noindent where $N \equiv 0 \pmod{D}$ is the total number of negative samples for each positive triple.

The micro-batch and negative sample structures used by BESS (\cref{eqn: microbatch,eqn: negsamples}) present three main advantages. Firstly, the fact that each micro-batch uses entities coming from all partitions $\mathcal{E}_j$, both for positive and negative triples, mitigates a potential source of bias and ensures a variety which is beneficial to the final embedding quality \cite{parallel_training_KGE}. Secondly, as $\mathcal{B}_i$ is processed on worker $i$ (which stores the embeddings for entities $\mathcal{E}_i$) only tail embeddings (positive and negative) need to be communicated between workers. Thirdly, by taking an equal number of triples from each partition $\mathcal{T}_{i,j}$ and of corrupted tails from each $\mathcal{E}_j$ we can efficiently organise the embedding sharing by means of AllToAll collectives, as every pair of workers needs to exchange the same amount of data. More specifically, the data sent from worker $j$ to worker $i$ consists of the embeddings of the $B/D$ tail entities in $\mathcal{B}_{i,j}$ and the $N/D$ entities in $\mathcal{N}_{i,j}$. This also implies that communication costs are constant across training steps and every worker performs the same amount of work, so that -- even with frequent synchronisations -- no significant idle time is introduced. 
 
\paragraph{Inference}

Different communication patterns are required at inference time, where a query $(h,r,?)$ needs to be scored against all tails $t \in \mathcal{E}$. A micro-batch of queries $\mathcal{Q}_i = \left\{ (h,r): h \in \mathcal{E}_i \right\}$ is fed by the master process to worker $i=1, \dots, D$. The relevant head entities are gathered from local memory and then shared through an AllGather collective between all devices. Worker $i$ proceeds to score the queries $(h, r) \in \bigcup_{j=1}^D \mathcal{Q}_j$ against all local tails $t \in \mathcal{E}_i$ and returns the top-$K$ predictions (with the corresponding scores) to the host, where for each query a final top-$K$ reduction is performed on the $D \cdot K$ retrieved scores in order to select the model's set of predictions. 

\subsection{Hardware Considerations}

Training performance depends directly on computation and communication costs, and indirectly on achievable batch size within a memory limit. Up to small constant relative factors, the time taken for computing a single training step is
\begin{align*}
t_{\textrm{compute}} &= c_{\textrm{compute}} \cdot (B \cdot N \cdot d + (B + N) \cdot |\boldsymbol{e}_F| \cdot d), \\
t_{\textrm{comms}} &= c_{\textrm{comms}} \cdot (B + N) \cdot (d + |\boldsymbol{e}_F|),
\end{align*}
with local memory usage
\begin{equation*}
    (B + N) \cdot (d + |\boldsymbol{e}_F|) < L,
\end{equation*}
where $c_{\textrm{compute}}$, $c_{\textrm{comms}}$ and $L$ are hardware-specific constants. If we assume $d$ and $|\boldsymbol{e}_F|$ are fixed, the amount of useful work done in a training step is proportional to $B \cdot N$. Efficient training therefore requires large $B$ and $N$, within the limit imposed by local memory. A hardware platform for efficient training requires low $c_{\textrm{compute}}$ or high achieved FLOP/s\footnote{FLOP/s: floating-point operations per second}. It also requires sufficiently low $c_{\textrm{comms}}$ or high memory bandwidth (byte/s), and high $L$ or large local memory (bytes), although these can be traded off against each other.

Our training system uses a single Bow~Pod$_{16}$, providing 16 IPUs each with 32~GiB streaming memory, 900~MiB in-processor memory and 350~TFLOP/s compute in FP16 precision \cite{bow_ipu}. IPUs are connected in a 2D torus by high-speed IPU-Links giving a total bidirectional bandwidth of 320~GiB/s between a chip and its peers. We designate each IPU as a worker ($D\!=\!16$) and reserve the entire streaming memory to store partitioned entity embeddings, associated optimiser state and features. To save memory and bandwidth these are stored in FP16. In-processor memory is used as a permanent store for all other parameters and optimiser state, for code and as working memory.

This configuration supports a maximum entity embedding size $d\!=\!416$, where for each entity its embedding, optimiser state and features are packed into a 4~kiB row in streaming memory. Maximum micro-batch size $B$ and negative sample size $N$ depend on scoring function and $d$, for example $\{\textrm{TransE}, d\!=\!256, B\!=\!512, N\!=\!1024\}$. In this example configuration, a single training step takes 6.5~ms, giving throughput $1.26 \cdot 10^6$~triples/s for an epoch time of 8~minutes. Inference uses a micro-batch size $|\mathcal{Q}_i|\!=\!128$ to compute top-$100$ predictions for the $\num{15000}$ validation samples in 102 seconds.

The Bow~Pod$_{16}$ hardware platform and Poplar software stack provide fine-grained control over on-device computation and access to streaming memory, enabling effective use of in-processor memory to achieve large batch size. This allows for reasonably efficient operation with the available memory communication bandwidth.

\section{Experimental Setup}
Models have been trained on a Bow~Pod$_{16}$ with a micro-batch size of 256-640 (per IPU) for $5\cdot 10^6$ steps, corresponding to 34-85 epochs. The MRR was evaluated periodically during training on $\num{15000}$ training samples and on the validation set. Hyperparameter settings for the different scoring functions can be found in \Cref{tab:model_params}\footnote{Detailed information on the hyperparameter settings and accuracy of all models used in the final ensemble can be found at \url{https://github.com/graphcore/distributed-kge-poplar/tree/resources/2022-ogb-submission}}. All scoring functions have been trained in combination with both log-sigmoid loss and sampled softmax cross entropy loss. TransE, TransH, and RotatE models have been trained with $L^1$ and $L^2$ distances. For the majority of models the learning rate has been decayed linearly to zero over the course of training.

\begin{table}[htb]
    \caption{Typical model configurations.}
    \begin{center}
    \begin{tabular}{ l l l l l l} 
    \toprule
    \textbf{Scoring function} & \textbf{Initial learning rate} & \textbf{Micro-batch size} & \textbf{Negative sample size} & \textbf{Embedding size} \\
    \midrule
    TransE & $[5\cdot 10^{-4}, 3\cdot 10^{-3}]$ & $\{256, 448, 512\}$ & $\{768, 1024, 1088, 1280\}$ & $\{256, 384\}$ \\
    TransH & $[2\cdot 10^{-4}, 4\cdot 10^{-3}]$ & $256$ & $256$ & $256$ \\
    RotatE & $[5\cdot 10^{-4}, 10^{-2}]$ & $\{256, 448\}$ & $\{256, 1088, 1280\}$ & $256$ \\
    DistMult & $[10^{-3}, 6\cdot 10^{-3}]$ & $\{256, 512\}$ & $\{1024, 1280\}$& $\{256, 384, 400\}$ \\
    ComplEx & $[5\cdot 10^{-4}, 5\cdot 10^{-3}]$ & $\{256, 512, 640\}$ & $\{896, 1024, 1280, 1536\}$& $400$ \\
    \bottomrule
    \end{tabular}
    \end{center}
    \label{tab:model_params}%
\end{table}

\section{Results}
\subsection{Individual Models}
\begin{figure}[tb]
    \centering
    \begin{subfigure}[t]{0.475\textwidth}
        \vskip 0pt
        \centering
        \includegraphics[height=4cm]{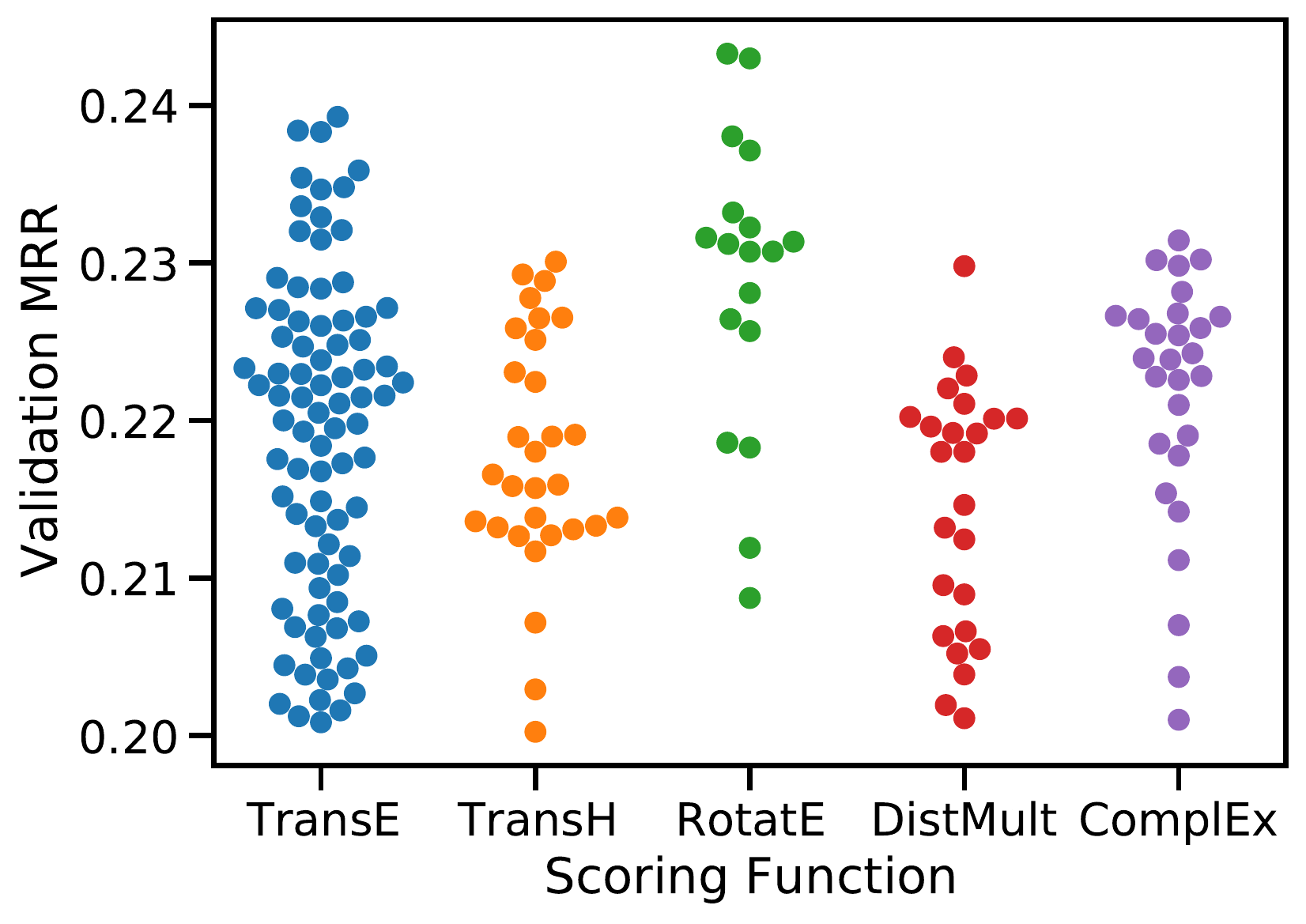}
        \subcaption{}
        \label{subfig:score_fn}
    \end{subfigure}
    \begin{subfigure}[t]{0.475\textwidth}
        \vskip 0pt
        \centering
        \includegraphics[height=4cm]{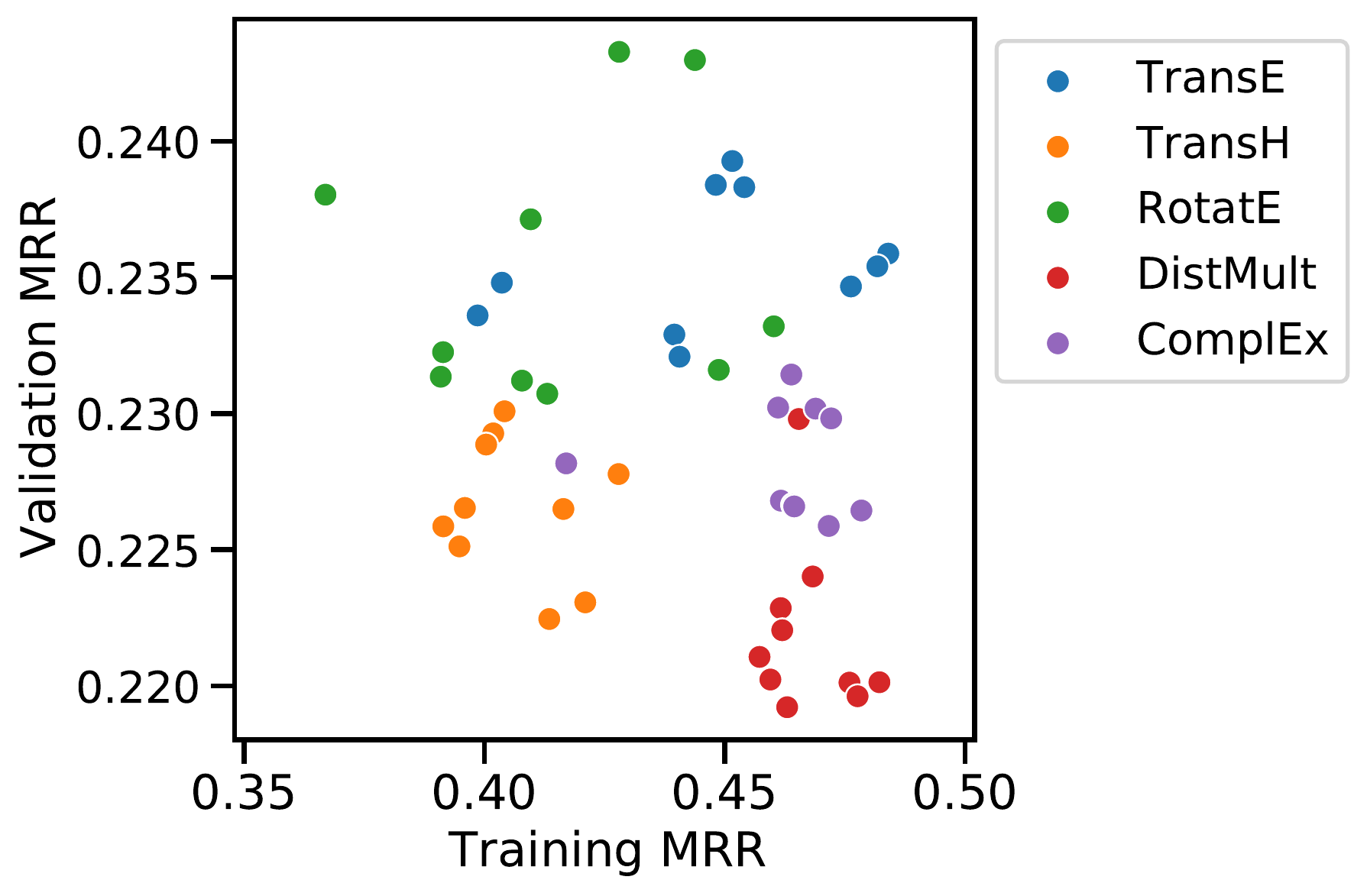}
        \subcaption{}
        \label{fig:valid_train}%
    \end{subfigure}
    \caption{ (a) Validation MRRs of the 185 models (84 TransE, 30 TransH, 28 ComplEx, 25 DistMult, 18 RotatE) that have been trained to a validation MRR of at least 0.2. (b) Validation MRR of the $10$ best models per scoring function plotted against their respective training MRR.}%
\end{figure}

To achieve the best possible MRR we aimed at maximising the diversity of models in our ensemble. This approach can be expected to benefit from complementary properties of different models, such as the properties of scoring functions specified in \Cref{table: decoders} and their different capabilities to model one-to-one or many-to-one relations. We trained a total of 259 models  to completion with different scoring functions (and distances), loss functions and sets of hyperparameters. Out of these models, 185 achieved a validation MRR > 0.2 (\Cref{subfig:score_fn}). 

Depending on the scoring function used by the model, a different tendency to overfit on the training data can be observed. In particular, models using DistMult or ComplEx reach a substantially higher MRR on the subsample of the training set than on the validation set (\Cref{fig:valid_train}). 

\begin{figure}[t]
    \centering
    \begin{subfigure}[t]{0.475\textwidth}
        \vskip 0pt
        \centering
        \includegraphics[height=4cm]{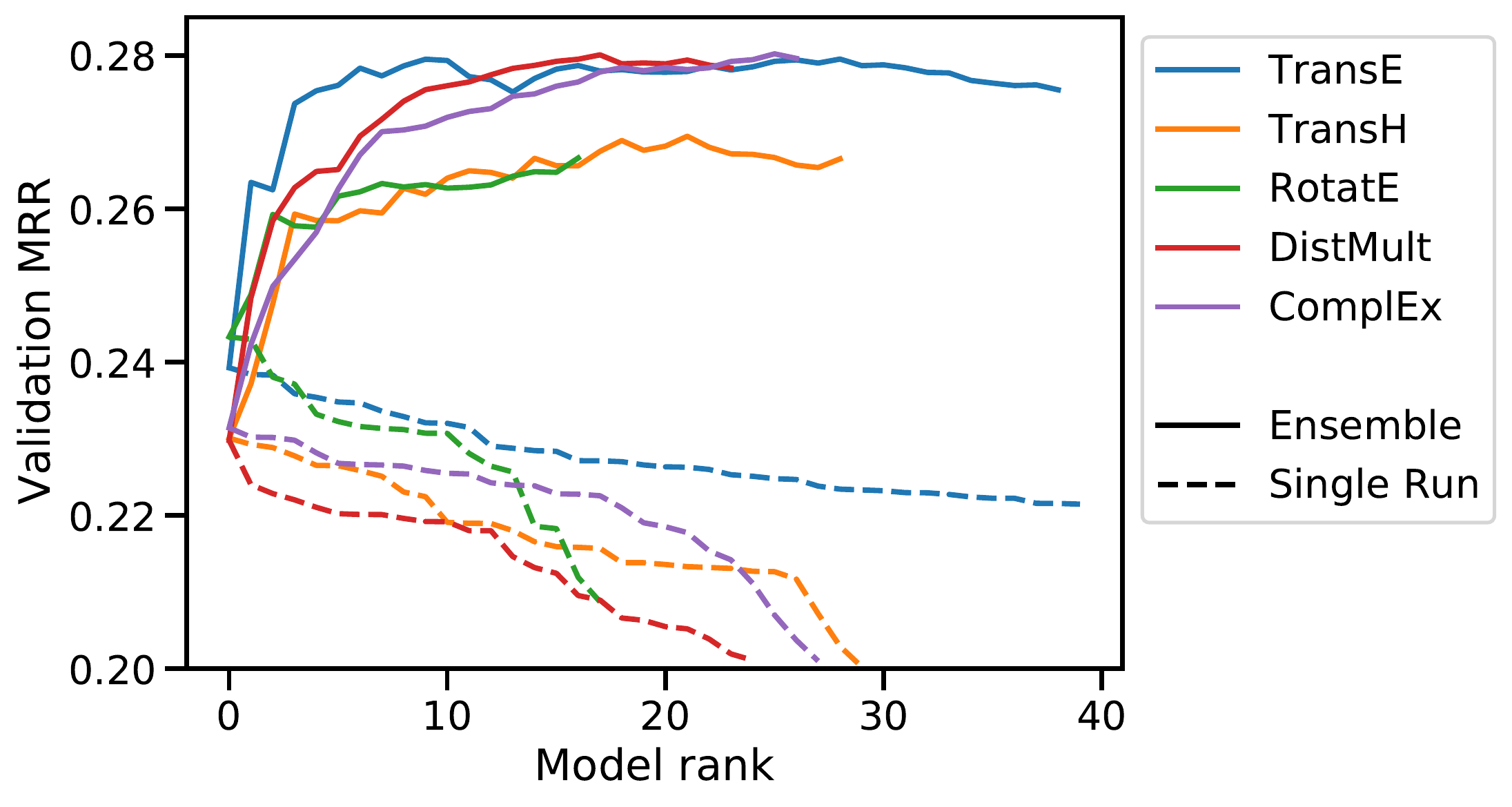}
        \subcaption{}
        \label{subfig:individual_ensemble}
    \end{subfigure}
    \begin{subfigure}[t]{0.475\textwidth}
        \vskip 0pt
        \centering
        \includegraphics[height=4cm]{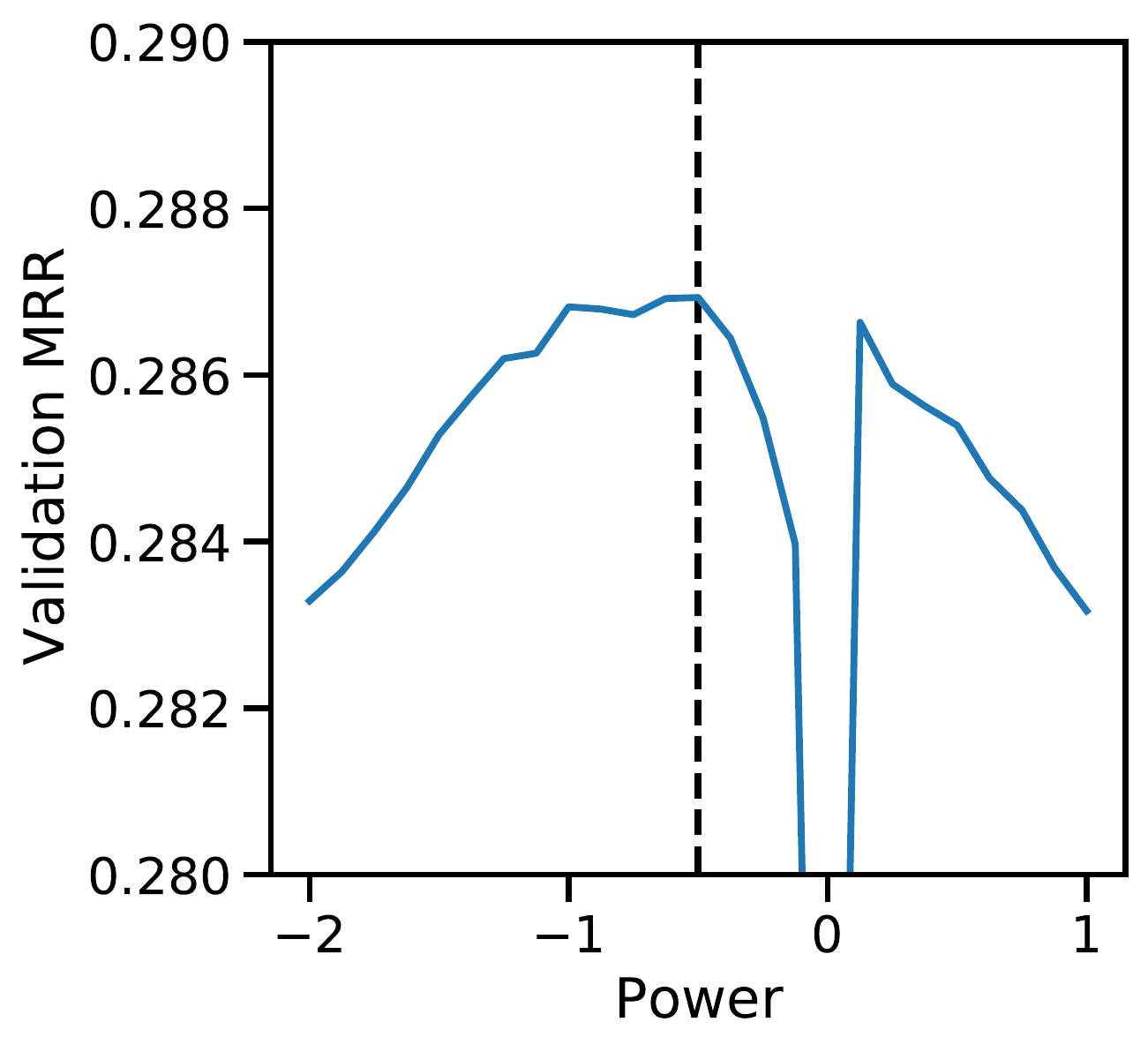}
        \subcaption{}
        \label{subfig:ensemble_power}
    \end{subfigure}
    \vskip\baselineskip
    \centering
    \begin{subfigure}[t]{0.475\textwidth}
        \vskip 0pt
        \centering
        \includegraphics[height=4cm]{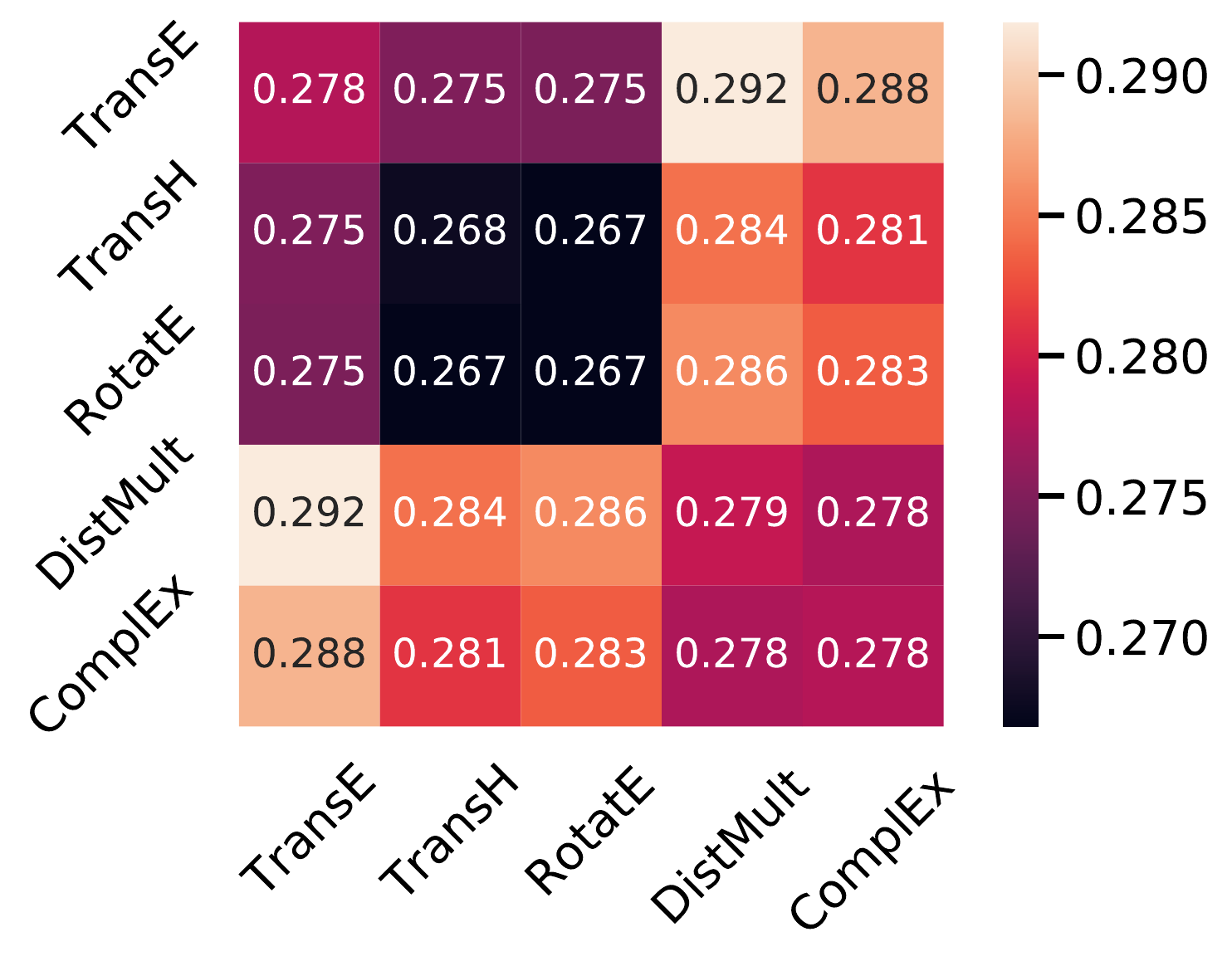}
        \subcaption{}
        \label{subfig:pairwise_ensemble}
    \end{subfigure}
    \begin{subfigure}[t]{0.475\textwidth}
        \vskip 0pt
        \centering
        \includegraphics[height=4cm]{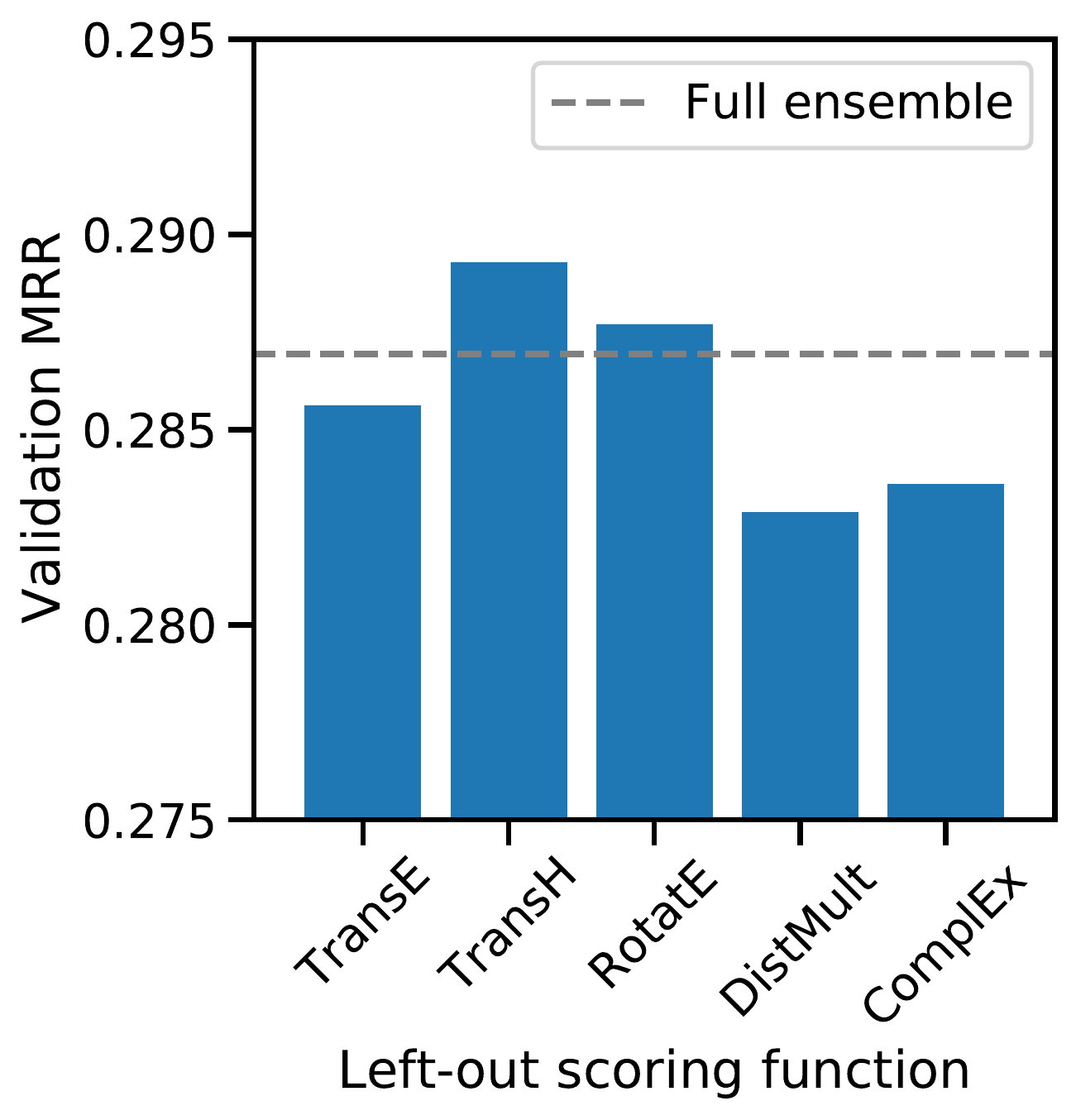}
        \subcaption{}
        \label{subfig:leave_out}
    \end{subfigure}
    \caption{(a) Validation MRR of the $k$-th best individual model (dashed) and of the ensemble of the $k$ best models (solid) per scoring function. (b) The effect of the power hyperparameter $p$ in \cref{eqn: ensemble} on validation MRR for an ensemble of $50$ models (the best $10$ for each scoring function). (c) Validation MRR of the ensemble of the $20$ best models per scoring function (main diagonal) and the $10+10$ best models of two scoring functions (off-diagonal). (d) Ablation of the contribution of different scoring functions to the ensemble of the $10$ best models for each scoring function.}
\end{figure}

\begin{figure}[htb]
    \centering
    \begin{subfigure}[b]{0.475\textwidth}
        \includegraphics[height=4cm]{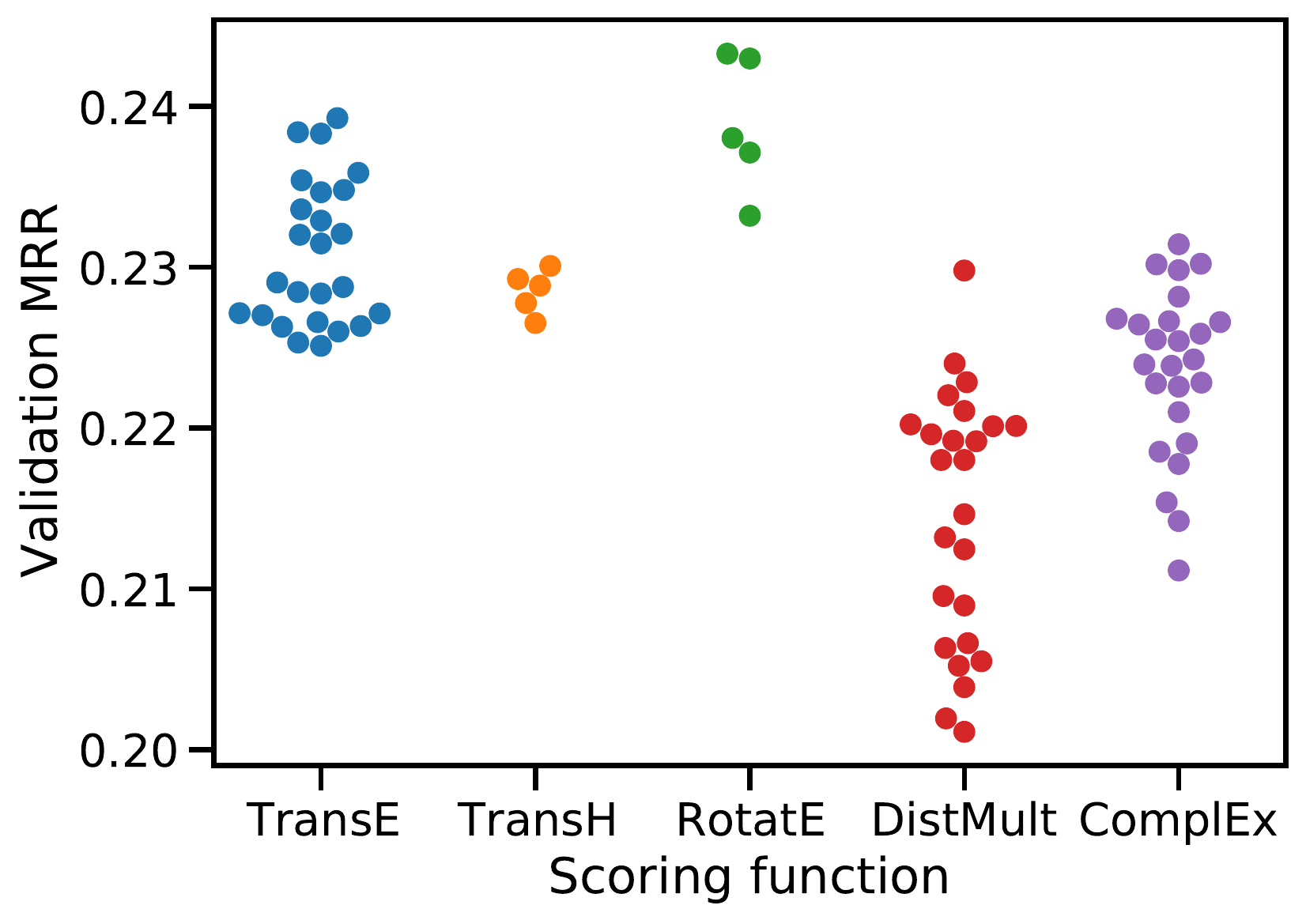}
        \caption{}
        \label{subfig:models_ensemble}
    \end{subfigure}
    \begin{subfigure}[b]{0.475\textwidth}
        \includegraphics[height=4cm]{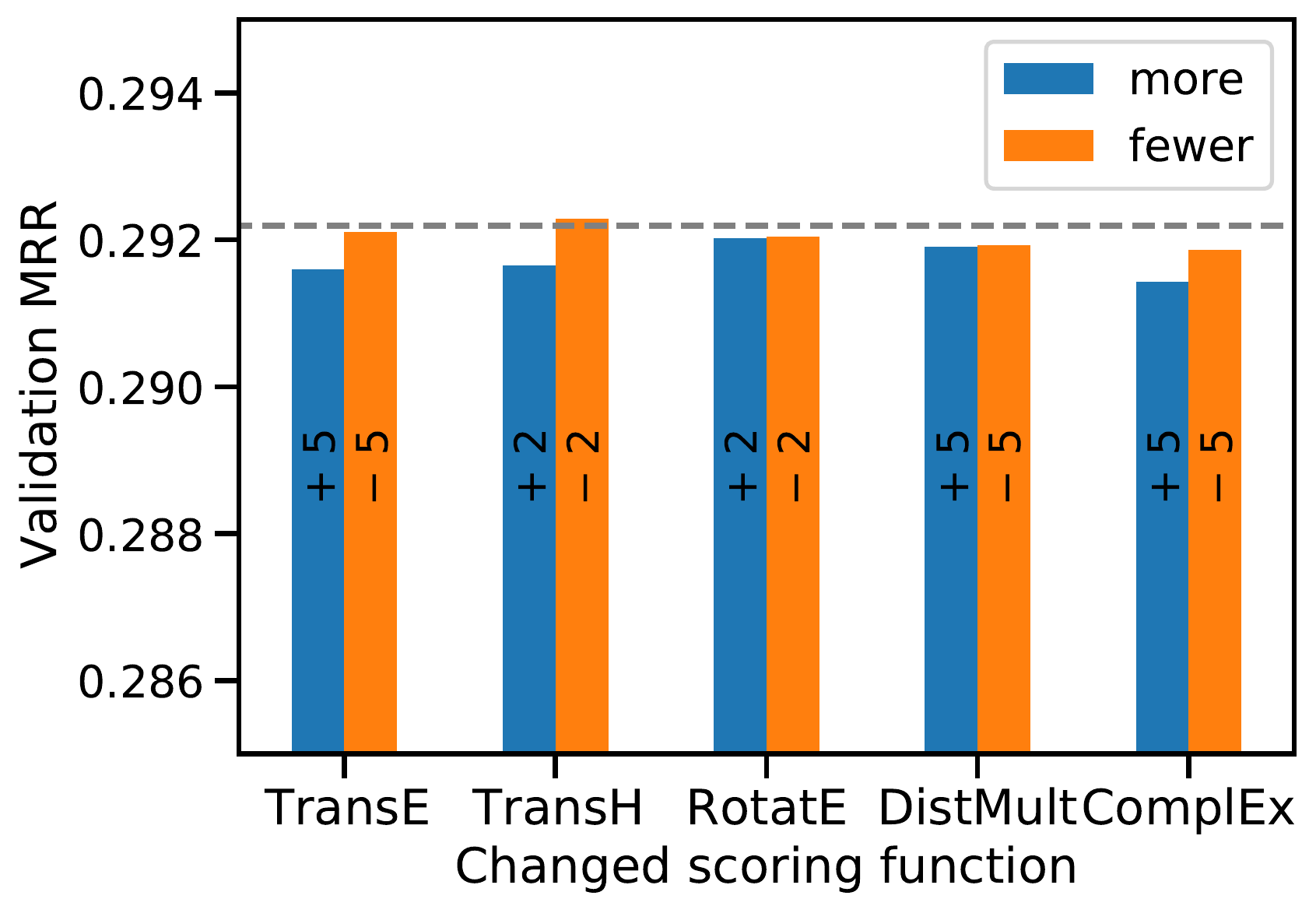}
        \caption{}
        \label{subfig:change_ensemble}
    \end{subfigure}
    \caption{(a) 85 models (25 TransE, 5 TransH, 5 RotatE, 25 DistMult, 25 ComplEx) have been selected for the final ensemble based on their validation MRR. (b) Increasing (blue) or reducing (orange) the number of models per scoring function does not substantially improve validation MRR of the ensemble.} \label{fig:final_ensemble}
\end{figure}

\subsection{Ensemble}
Using the mean-ensembling strategy laid out in \cref{eqn: ensemble} to create an ensemble, powers $p\in[-1.0, -0.5]$ yield good results (\Cref{subfig:ensemble_power}). As relying less on few top results intuitively generalises better, we selected $p=-0.5$ for the final ensemble.

Depending on the scoring function, models benefit to a different degree from ensembling: although the best individual models use RotatE (\Cref{subfig:score_fn}), ensembles of a single scoring function among TransE, DistMult and ComplEx yield a higher validation MRR than ensembles of TransH or RotatE models (\Cref{subfig:individual_ensemble}). When ensembling two different scoring functions, the best results are achieved by including DistMult or ComplEx (\Cref{subfig:pairwise_ensemble}). Likewise, removing models using DistMult or ComplEx from an ensemble results in a substantial MRR degradation, while leaving out models with TransH or RotatE can even be beneficial (\Cref{subfig:leave_out}). A possible explanation for these observations can be found in the high training MRR achieved by DistMult and ComplEx models (\Cref{fig:valid_train}), which produces good generalisation when these models' tendency to overfit is mitigated by the regularising effect of mean-ensembling.

Based on this evidence, individual models have been ranked by validation MRR and a diverse ensemble consisting of $85$ models (the $25$ best TransE, DistMult and ComplEx models, and the $5$ best TransH and RotatE models; \Cref{subfig:models_ensemble}) was selected, achieving a validation MRR of 0.2922 and an MRR of 0.2562 on the test-challenge set. Changing the composition of this ensemble did not further improve validation MRR (\Cref{subfig:change_ensemble}).


\section{Conclusions}
We demonstrate the distributed training of large Knowledge Graph Embedding models on a Graphcore Bow~Pod$_{16}$ system. Enabled by the fast execution scheme of the distribution framework BESS, we show the substantial advantage of large ensembles of a diverse set of models over individual KGE models. With an MRR of 0.2562 on the test-challenge set, the solution laid out in this paper has achieved first place in the WikiKG90Mv2 track of the Open Graph Benchmark Large-Scale Challenge at NeurIPS 2022 \cite{ogb_results}.

\section*{Acknowledgements}
We thank Luke Hudlass-Galley for his helpful comments on the manuscript. We are grateful for all the support received from our Graphcore colleagues.

\bibliographystyle{apacite}
\bibliography{KGbib}

\end{document}